\pdfoutput=1

\documentclass[10pt, a4paper, twocolumn]{naverlabseurope}

\usepackage{times}
\usepackage{latexsym}
\usepackage{booktabs}
\usepackage{graphicx}
\usepackage{enumitem}
\usepackage{amsmath,graphicx}
\usepackage{dblfloatfix}
\usepackage{amsmath,bbm}
\usepackage{hyperref}
\usepackage[T1]{fontenc}

\usepackage[utf8]{inputenc}

\usepackage{microtype}

\usepackage{inconsolata}

\newcounter{mycounter}

\title{An Adapter-Based Unified Model for Multiple Spoken Language Processing Tasks}

\authors{Varsha Suresh\textsuperscript{1}, Salah Aït-Mokhtar\textsuperscript{2}, Caroline Brun\textsuperscript{2}, Ioan Calapodescu\textsuperscript{2}}

\affiliations{\textsuperscript{1}National University of Singapore, \textsuperscript{2}Naver Labs Europe}
\website{}
\websiteref{}
\contributions{}
\teasercaption{}

\begin{abstract}
Self-supervised learning models have revolutionized the field of speech processing. However, the process of fine-tuning these models on downstream tasks requires substantial computational resources, particularly when dealing with multiple speech-processing tasks. In this paper, we explore the potential of adapter-based fine-tuning in developing a unified model capable of effectively handling multiple spoken language processing tasks. The tasks we investigate are Automatic Speech Recognition, Phoneme Recognition, Intent Classification, Slot Filling, and Spoken Emotion Recognition. We validate our approach through a series of experiments on the SUPERB benchmark, and our results indicate that adapter-based fine-tuning enables a single encoder-decoder model to perform multiple speech processing tasks with an average improvement of 18.4~\% across the five target tasks while staying efficient in terms of parameter updates.
\end{abstract}

\begin{document}
\maketitle
\section{Introduction}
The fine-tuning of self-supervised learning (SSL) models, such as wav2vec~2.0 \cite{baevski2020wav2vec}, has improved the performance of Spoken Language Processing (SLP) tasks. However, as the quality of representations generated by these models improves, there is a corresponding increase in their size, necessitating additional storage and computational resources. This issue becomes particularly pronounced when dealing with multiple speech-processing tasks, with each target task requiring separate model fine-tuning, further increasing the need for resources.

Modular architectures, such as adapters, have been widely used in NLP to tackle both parameter efficiency and multi-tasking \cite{houlsby2019parameter}. While adapter-based fine-tuning has been utilized in speech-related tasks, such as speech translation \cite{le2021lightweight,gow2023naver,antonios2022findings} and domain adaptation \cite{thomas2022efficient}, its efficiency in developing a unified model capable of handling multiple Spoken Language Processing (SLP) tasks remains relatively unexplored. Existing attempts to model multiple SLP tasks with a single model utilises task-specific decoders \cite{yang2021superb}. However, this approach becomes less scalable as the number of tasks increases.

In this work, we aim to develop a scalable and parameter-efficient unified encoder-decoder model to effectively handle multiple spoken language processing (SLP) tasks. For this, we use adapters \cite{houlsby2019parameter}, which allows new tasks to be added without the need to re-train the entire model and which also mitigates the need for dedicated decoders \cite{yang2021superb}. Moreover, since adapters facilitate Multi-Task Learning (MTL), we investigate two approaches: Stacking \cite{pfeiffer2020adapterhub} and Fusion \cite{zhao2022multimodal}, in addition to single-task adapters. To evaluate our approach, we choose five speech-processing tasks from the SUPERB benchmark \cite{yang2021superb}: Automatic Speech Recognition (ASR), Phoneme Recognition (PR), Intent Classification (IC), Slot Filing (SF), and Spoken Emotion Recognition (ER). The detailed model description is provided in Figure \ref{fig:model}. From our experiments, we observed that adapter-based fine-tuning outperformed the SUPERB benchmark with an average improvement of 18.4~\% achieved across 5 target tasks. We summarise our contributions below:

\begin{figure*}[t]
    \centering
    \includegraphics[width=0.9\linewidth]{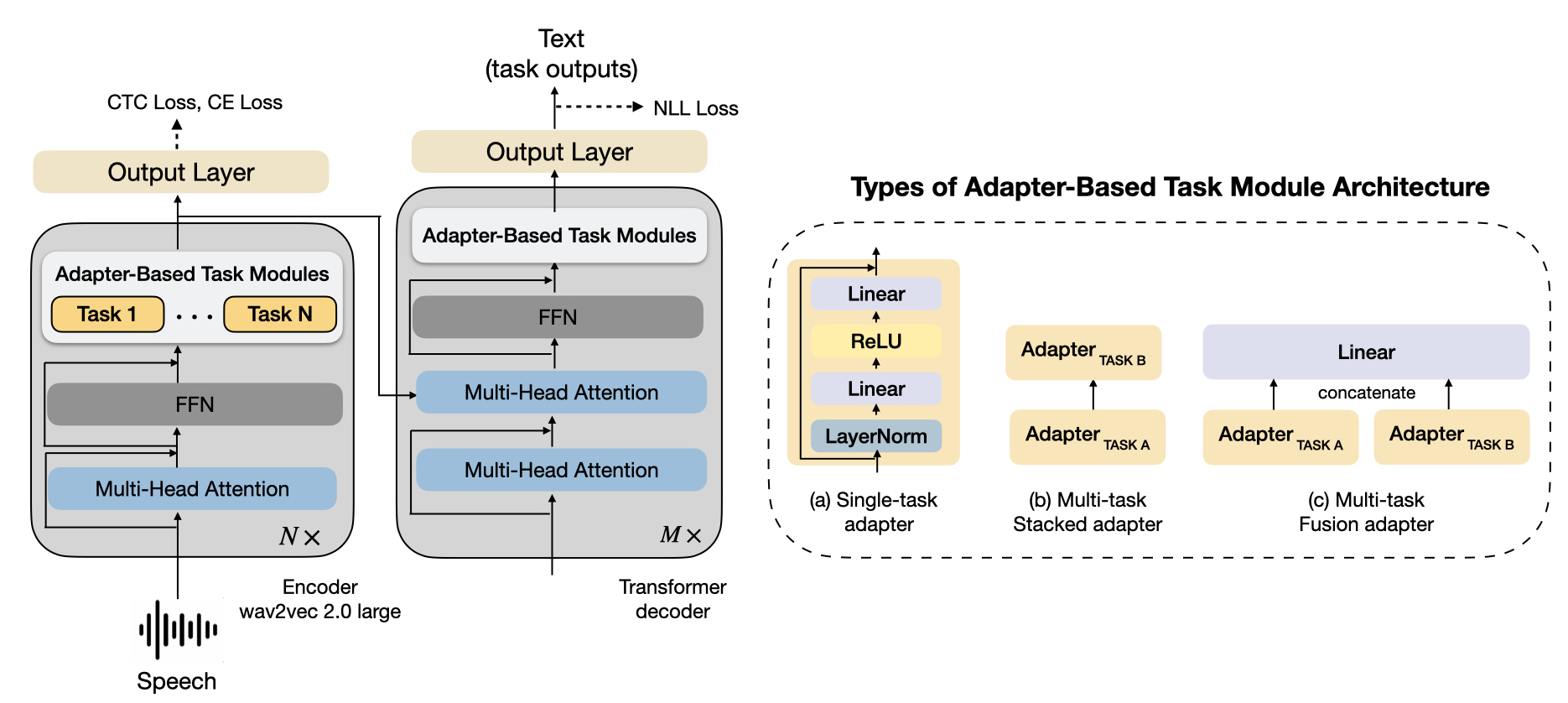}
  \caption{\textbf{Left}: Overall model architecture with a unified encoder-decoder model with adapter-based task modules on each transformer layer. \textbf{Right}: Three types of adapter-based task modules used.}
  \label{fig:model} 
\end{figure*}

\begin{itemize}
    \item We investigate the feasibility and efficiency of using adapters to build a unified encoder-decoder model that can tackle multiple spoken language processing tasks in a simple and scalable manner.  
    \item We explore multi-task learning within our unified framework with two methods: stacking and fusion, which combine adapters to enhance the performance of positively correlated tasks.
\end{itemize}
\section{Related Work}

In the field of NLP, researchers have used a single model to handle multiple tasks and adapt them to different domains \cite{mccann2018natural}. In the speech domain, most approaches that deal with multiple tasks fall under multi-task models. They either focus on improving a primary task by using auxiliary tasks, like performing ASR to enhance Emotion Recognition \cite{cai2021speech,li2022fusing,feng2020end}, or simultaneously perform multiple tasks ---slot-filling and intent classification \cite{li2018joint}, ASR and speech translation 
\cite{tang2021general,radford2022robust} etc. However, these approaches are not easily scalable for new tasks and are mostly applied for tasks that are known to be positively correlated.

Some studies have aimed to create a unified model for multiple speech-processing tasks by training different modules and composing them to perform each task. These architectures comprise encoders and decoders trained to capture features from different modalities, such as text and speech \cite{ao2022speecht5}, or speech characteristics like prosody \cite{chen2021speechnet} etc.  In contrast, our approach focuses on constructing task-specific modular architectures. Furthermore, adding a task is straightforward as these task-specific modules are trained independently.  

Our work is inspired by SUPERB \cite{yang2021superb}, where multiple tasks are modeled using pre-trained frozen encoders (such as wav2vec~2.0) and task-specific decoders, the task performance relying on the type of decoder used for the task \cite{zaiem2023speech}. However, this approach does not scale well as the number of tasks increases. Instead, in our work, we aim to develop a single encoder-decoder model and show that adapters on the decoder side help us adapt to different types of tasks (both classification and generative) without the need for dedicated decoders.

\section{Model Architecture}

\subsection{Pre-trained Encoder-Decoder model}
We use wav2vec~2.0-large\footnote{\url{https://huggingface.co/facebook/wav2vec2-large-lv60}} as encoder and a 6-layer transformer decoder which is randomly initialized. We fine-tuned this encoder-decoder model on the LibriSpeech 100-hr dataset \cite{panayotov2015librispeech} for the ASR task using hybrid CTC/attention objective \cite{watanabe2017hybrid} as the base model for our unified model. We use SentencePiece (BPE) vocabulary of size 5000. This base model achieved a word error rate (WER) of 3.54 on the test-clean split of LibriSpeech which is comparable to the 3.1 WER reported in the SUPERB benchmark.
\begin{table*}[h]
\centering
\resizebox{0.9\linewidth}{!}{%
\begin{tabular}{l|cc|c|cc|c|c}
\hline
\multicolumn{1}{c|}{} & \multicolumn{2}{|c|}{LibriSpeech} & IEMOCAP & \multicolumn{2}{|c|}{SNIPS} & FSC \\
\multicolumn{1}{c|}{} & \begin{tabular}[c]{@{}c@{}}ASR\\ (WER $\downarrow$)\end{tabular} & \begin{tabular}[c]{@{}c@{}}PR\\ (PER $\downarrow$)\end{tabular} & \begin{tabular}[c]{@{}c@{}}ER\\ (Acc \% $\uparrow$)\end{tabular} & \begin{tabular}[c]{@{}c@{}}IC\\ (Acc \% $\uparrow$)\end{tabular} & \begin{tabular}[c]{@{}c@{}}SF\\ (F1 $\uparrow$,CER $\downarrow$)\end{tabular} & \begin{tabular}[c]{@{}c@{}}IC\\ (Acc \% $\uparrow$)\end{tabular} & Avg \\ \hline
WavLM large SUPERB \cite{yang2021superb} & 3.4 & 3.1  & \textbf{70.6} & - & 92.2, 18.4 & 99.0 & 89.5  \\ \hline
wav2vec2.0 large SUPERB \cite{yang2021superb} & \textbf{3.1} & 4.7  & 65.6 & - & 87.1, 27.3 & 95.2 & 85.5 \\ \hline
wav2vec2.0 large (Ours) & 3.5 & \textbf{2.4} & 68.2 & \textbf{99.1} & \textbf{95.4, 11.8} & \textbf{99.5} & \textbf{90.9}
\\ \hline
\end{tabular}%
}
\caption{Performance comparison in various speech processing tasks from the SUPERB benchmark. WavLM is currently ranked first in SUPERB's leaderboard and we choose wav2vec2-large to compare multi-decoder (SUPERB) and single decoder (Ours) solutions. Models from SUPERB have different decoder implementations for each task (e.g. Bi-LSTM, CNNs, linear projections) on top of the chosen SSL model. Our approach is a single transformer encoder-decoder model capable of performing all six tasks using various adapters for each task and initialized on the encoder side with the chosen SSL model. The metrics are computed with the s3prl framework and Avg denotes the average performance across all the tasks.}
\label{tab:overall}
\end{table*}
\begin{table*}[h]
\centering
\resizebox{\textwidth}{!}{%
\begin{tabular}{l|cc|ccc|cc}
\hline
 & \multicolumn{2}{c|}{IEMOCAP} & \multicolumn{3}{c|}{SNIPS} & \multicolumn{2}{c}{FSC} \\
 & \begin{tabular}[c]{@{}c@{}}ASR\\ (WER $\downarrow$)\end{tabular} & \begin{tabular}[c]{@{}c@{}}ER\\ (Acc \% $\uparrow$)\end{tabular} & \begin{tabular}[c]{@{}c@{}}ASR\\ (WER $\downarrow$)\end{tabular} & \begin{tabular}[c]{@{}c@{}}IC\\ (Acc \% $\uparrow$)\end{tabular} & \begin{tabular}[c]{@{}c@{}}SF\\ (F1 $\uparrow$,CER $\downarrow$)\end{tabular} & \begin{tabular}[c]{@{}c@{}}ASR\\ (WER $\downarrow$)\end{tabular} & \begin{tabular}[c]{@{}c@{}}IC\\ (Acc \% $\uparrow$)\end{tabular} \\ \hline
MTL: ESP-net \cite{arora2022espnet} & - & 67.6 & - & 91.7 & - & - & \textbf{99.6} \\ 
MTL: ASR+SER \cite{li2022fusing} & 32.7 & $\text{63.4}^*$ & - & - & - & - & - \\
MTL: ASR+IC \cite{meeus2022multitask} & - & - & - & - & - & - & 98.2 \\
MTL: ASR+IC \cite{lai2020towards}& - & - & 11.8 & 98.6 & - & - & - \\
\hline
wav2vec2.0 large (Ours) &  &  &  &  &  &  &  \\
- Single task Adapter & 22.3 & 65.6 & 8.5 & 98.4 & 94.7, 12.9 & 0.6 & 99.4 \\
- MTL: Stacked & 24.2 & \textbf{68.2} & 7.7 & 98.7 & 94.4, 13.5 & 0.6 & 99.5 \\
- MTL: Fusion & \textbf{22.1} & 65.4 & \textbf{7.3} & \textbf{99.1} & \textbf{95.4, 11.8} & 0.6 & 99.3 \\ \hline
\end{tabular}%
}
\caption{Performance comparison between various MTL implementations and our three different adapter-based architectures. *uses weighted accuracy}
\label{tab:ablation}
\end{table*}
\begin{table}[]
\centering
\begin{tabular}{l|cc}
\hline
\multicolumn{1}{c|}{} & \multicolumn{2}{c}{\# of tasks}         \\
\multicolumn{1}{c|}{} & 6 tasks & 9 tasks\\ \hline
SUPERB \cite{yang2021superb} & 126.6M        & 252.8M \\
Ours & \textbf{113.1M}      & \textbf{135.6M}      \\ \hline
Ratio & 89.3\%  & 53.6\%    \\ \hline
\end{tabular}
\caption{Comparison between the total \# of additional trainable parameters required to accommodate 6 tasks depicted in Table \ref{tab:overall} and 9 tasks which includes the additional ASR tasks in Table \ref{tab:ablation}.}
\label{tab:param}
\end{table}

\subsection{Adapter-based Task Modules}
To enable the above-mentioned pre-trained encoder-decoder model to perform multiple SLP tasks, we insert task-specific adapter modules into the transformer \cite{vaswani2017attention} layers of both the encoder and the decoder. These modules are depicted in Figure~\ref{fig:model}. The remainder of the model is frozen.

We focus on three types of architecture: i) single adapter ii) adapter stacking, and iii) adapter fusion as shown on the right side of Figure~\ref{fig:model}. In the standard setting, a single adapter is trained for each task \cite{houlsby2019parameter}. However, some SLP tasks are known to benefit from MTL, such as performing ASR and emotion recognition \cite{li2022fusing,feng2020end} and Intent classification and Slot-filling \cite{weld2022survey}. As adapters naturally support MTL \cite{houlsby2019parameter} in addition to using a single adapter per task, we use the adapter stacking \cite{pfeiffer2020adapterhub} and adapter fusion settings (same as the fast fusion setting in \cite{zhao2022multimodal}) to perform positively correlated tasks together.

 To facilitate a unified model, we encompass both classification (e.g., emotion recognition) and generative (e.g., slot filling) SLP tasks into this single encoder-decoder model. To achieve this, we model the classification task as a generative task i.e., the classification labels are generated. To accommodate multiple tasks using a single decoder output, some task-specific tokens are allocated in the vocabulary---slot value for SF task, emotion labels for ER task, etc. These tokens are selected from the least frequently used tokens in the vocabulary. For MTL, we combine the ground truth of the tasks involved using a task separator token. For example, to perform ASR along with ER, we format the ground truth as $<$transcript$>$ $<$task separator$>$ $<$emotion label$>$. 

For training the encoder-decoder model, we use a combination of losses depending on the adapter architecture and the task. The overall objective $\mathbf{L}$ can be written as,
\begin{gather}
L_{nll} = \sum_{task = 1}^{N} \lambda_{task}  \cdot L_{task}\\
\mathbf{L} = (1-\lambda_{ctc}) \cdot L_{nll} + \lambda_{ctc} \cdot L_{ctc} + \mathbbm{1}_{ce} \cdot L_{ce}   
\end{gather}
where $L_{nll}$ denotes the
Negative Log-Likelihood loss at the decoder end. The output tokens during multi-task training comprise tokens from $N$ tasks which are weighted using the hyperparameter
$\lambda_{task}$. $L_{ctc}$ denotes CTC loss applied at the encoder end, similar to hybrid CTC/attention objective \cite{watanabe2017hybrid}. Hyperparameter $\lambda_{ctc}$ is used to weigh between the NLL and CTC loss. Finally, $L_{ce}$ denotes Cross Entropy Loss applied at the encoder end for classification tasks, and
hyperparameter $\mathbbm{1}_{ce}$ is 1 when it's a classification task. 
 
\section{Experimental Setup}

We train adapters to perform five different SLP tasks (corresponding datasets are denoted in the brackets), 1) ASR (LibriSpeech \cite{panayotov2015librispeech}), 2) PR (LibriSpeech), 3) ER (IEMOCAP \cite{busso2008iemocap}), 4) IC (Fluent Speech Commands \cite{lugosch2019speech}), and 5) SF (SNIPS \cite{DBLP:journals/corr/abs-1805-10190}). We chose datasets used by the SUPERB benchmark\footnote{\url{https://github.com/s3prl/s3prl}} \cite{yang2021superb} for the corresponding tasks. In addition, we also train ASR adapters specifically for each domain (IEMOCAP, SNIPS and FSC) which helps in MTL (e.g, ASR + ER) and IC adapter for SNIPS which helps when performed with SF.

For evaluation, we follow the same setting as SUPERB. The adapter dimension was set to 128. For $\lambda$ settings, in single-label classification tasks such as ER, $\lambda_{ctc}$ is set to 0. For the rest, $\lambda_{ctc}$ is set to 0.3 and in experiments where CTC loss was used, we combined the attention-based and CTC scores for joint decoding, assigning a weight of 0.4 to the CTC scores (following the SpeechBrain recipe). For MTL, in the stacked adapter setting \cite{pfeiffer2020adapterhub} only the additional adapter is trained, while the rest of the model, including the bottom adapter(s) remains frozen. Here, $\lambda_{task}$ assigns a higher weight to the tokens corresponding to the new task. In our experiments, this value was set to 0.9 for the new task and 0.1 for the tasks of the already-trained bottom adapters. In fusion, the adapters are already trained with respective tasks, so we experimented with two settings: first, $\lambda_{task}$ is set to 1, and second, we set it to equal weights for all tasks and chose the best. We modified SpeechBrain \footnote{\url{https://github.com/speechbrain/speechbrain}} recipes for our implementation. 

\section{Results and Discussion}

Table \ref{tab:overall} presents our results, showing that our approach achieves better performance compared to the wav2vec2 SUPERB \cite{yang2021superb} benchmark (+5.4) and actually also with the WavLM model (+1.4) also from SUPERB. This performance improvement can be attributed to our design choice of utilizing adapters that allows combining different tasks for improved multi-task learning. For example, for SF performance on SNIPS, the adapters where ASR, SF and IC are learned simultaneously allows an improvement of 8.3 F1 and 15.5 CER (see Table \ref{tab:ablation}). Detailed results regarding the performance of different adapter combinations are discussed in the Ablation Study. In contrast to having a frozen encoder and task-specific decoders, we incorporate task-specific adapters on a single encoder-decoder model to perform multiple SLP tasks which leads to both efficient utilization of encoder representations, and memory efficiency (see Table \ref{tab:param}). 

\subsection{Ablation Study: Comparison between different types of adapter-based task modules}

Research has shown that certain tasks, like ASR and ER \cite{li2022fusing,feng2020end}, can benefit from simultaneous learning, enhancing each other's performance. As adapters naturally enable MTL \cite{houlsby2019parameter}, in addition to single-adapter task modules, we investigate two adapter-based MTL approaches: Stacking and Fusion. We hypothesize that performing MTL with adapters produces less increase in computational overhead compared to the performance improvement.\footnote{Adapter stacking: no change in the number of parameters and adapter Fusion introduces an additional 57M parameters.}

Table \ref{tab:ablation} compares the performance amongst three different adapter settings and also with existing works that perform MTL. In the IEMOCAP dataset, the Adapter Stacking setting achieves the highest performance in Emotion Recognition. On the SNIPS dataset, the Adapter Fusion setting performs the best in SF and IC. Our performance is comparable to studies that use gold-text directly, such as \cite{qin2021co} with an SF-F1 score of 95.9 and IC-Acc of 98.8\% on SNIPS. For FSC, there is minimal performance variation in the literature, since models already achieve above 99\% accuracy. The WER is also comparable with existing works --- Ours:~0.6, and \cite{fu2022multi}:~0.5. This performance improvement of adapter-based MTL architectures aligns with previous research indicating that MTL enhances task performance \cite{qin2021co}. Furthermore, fine-tuning our ASR adapters for each dataset performs better than approaches that use generic ASR models, as previously demonstrated by \cite{li2022fusing}. 

In addition to the improvements in performance, our unified model shows multi-task capabilities in a parameter-efficient and scalable manner. Table \ref{tab:param} illustrates the comparison between our approach and the SUPERB benchmark in terms of the number of trainable parameters needed for accommodating six tasks (as in Table \ref{tab:overall}) and nine tasks (including the additional ASR tasks from Table \ref{tab:ablation}). Notably, our approach requires fewer parameters, and more importantly, even as the number of tasks increases, the increase in the parameter count remains significantly lower with the ratio dropping to 53.6\%.

\section{Conclusion}

Our work shows that adapter-based task modules effectively enable a unified encoder-decoder model for handling multiple speech-processing tasks. Our experiments show that we are able to achieve performance improvements compared to the SUPERB benchmark, while being more efficient in terms of parameters by eliminating the need for dedicated task-specific decoders. This work highlights the potential to develop simple and scalable model architectures that are capable of performing multiple SLP tasks within a unified model. In the future, our goals include evaluating our approach for different choices of SSL models such as HuBERT and WavLM and exploring different adapter architectures. Additionally, we also aim to broaden the scope of our approach to add the remaining tasks in the  SUPERB benchmark such as Speaker Identification, Speaker Diarization, and other speech-processing tasks/datasets.

\bibliography{custom}
\bibliographystyle{ieeenat_fullname}

\end{document}